\documentclass[conference]{IEEEtran}
\IEEEoverridecommandlockouts
\usepackage{cite}
\usepackage{amsmath,amssymb,amsfonts}
\usepackage{algorithmic}
\usepackage{graphicx}
\usepackage{textcomp}
\usepackage{xcolor}
\usepackage{hyperref}
\usepackage{multirow}
\usepackage[flushleft]{threeparttable}
\def\BibTeX{{\rm B\kern-.05em{\sc i\kern-.025em b}\kern-.08em
    T\kern-.1667em\lower.7ex\hbox{E}\kern-.125emX}}
\begin{document}

\newcommand\blfootnote[1]{%
  \begingroup
  \renewcommand\thefootnote{}\footnote{#1}%
  \addtocounter{footnote}{-1}%
  \endgroup
}

\title{Attention-based Multi-Input Deep Learning Architecture for Biological Activity Prediction: \\An Application in EGFR Inhibitors
}
 
\author{\IEEEauthorblockN{Huy Ngoc Pham}
\IEEEauthorblockA{\textit{Research \& Development} \\
\textit{OPC Pharmaceutical Company}\\
Ho Chi Minh City, Vietnam \\
ngochuy.yds@gmail.com}
\and
\IEEEauthorblockN{Trung Hoang Le}
\IEEEauthorblockA{\textit{Research Engineer} \\
\textit{Trusting Social}\\
Ho Chi Minh City, Vietnam \\
le.hg.trung@gmail.com}}

% \author{
%     \IEEEauthorblockN{Huy Ngoc Pham \\ Ho Chi Minh City, Vietnam \\ \url{ngochuy.yds@gmail.com}}
%     \and
%     \IEEEauthorblockN{Trung Hoang Le \\ Ho Chi Minh City, Vietnam \\ \url{le.hg.trung@gmail.com}}
%     }

\IEEEoverridecommandlockouts
\IEEEpubid{\makebox[\columnwidth]{978-1-7281-3003-3/19/\$31.00 \copyright2019 IEEE \hfill} \hspace{\columnsep}\makebox[\columnwidth]{ }}

\maketitle

\IEEEpubidadjcol

\begin{abstract}
Machine learning and deep learning have gained popularity and achieved immense success in Drug discovery in recent decades. Historically, machine learning and deep learning models were trained on either structural data or chemical properties by separated model. In this study, we proposed an architecture training simultaneously both type of data in order to improve the overall performance. Given the molecular structure in the form of SMILES notation and their label, we generated the SMILES-based feature matrix and molecular descriptors. These data were trained on a deep learning model which was also integrated with the Attention mechanism to facilitate training and interpreting. Experiments showed that our model could raise the performance of prediction comparing to the reference. With the maximum MCC 0.58 and AUC 90\% by cross-validation on EGFR inhibitors dataset, our architecture was outperforming the referring model. We also successfully integrated Attention mechanism into our model, which helped to interpret the contribution of chemical structures on bioactivity.
% \footnote{Our models are available at \href{https://github.com/lehgtrung/egfr-att}{https://github.com/lehgtrung/egfr-att}}
\end{abstract}


%\blfootnote{Our models are available at \href{https://github.com/lehgtrung/egfr-att}{https://github.com/lehgtrung/egfr-att}}

\section{Introduction}
% Machine learning
Machine learning was applied widely in drug discovery, especially in virtual screening for hit identification. The most popular techniques are Support Vector Machine (SVM), Decision Tree (DT), k-Nearest Neighbor (k-NN), Naive Bayesian method (NB), and Artificial Neural network (ANN). \cite{Lavecchia2015}. In these methods, ANNs need not assume that there was any type of relationship between activity and molecular descriptors, and ANNs are usually outperforming in traditional Quantitative structure -- activity relationship problem because they can deal with both nonlinear and linear relationship. As a result, ANNs rose to become a robust tool for Drug Discovery and Development \cite{Winkler2004}. However, ANNs are usually sensitive to overfitting and difficult to design an optimal model. Additionally, ANNs also require huge computation resources and their results usually are unable to be interpreted. Those weaknesses could be a reason for limited use of neural network comparing to Decision Tree or Naive Bayesian algorithms \cite{Winkler2004, Lavecchia2015}.

Above algorithms can be applied to various types of chemical features. These features are either structural information or chemical properties. The structural information could be represented as fingerprint vector by using specific algorithms (e.g, Extended-Connectivity Fingerprints \cite{Rogers2010}, Chemical Hashed Fingerprint \cite{Al-Lazikani2004}) while chemistry information could be described by various molecular descriptors (e.g, logP, dipole moment). The ideas that combine some types of features to improve the overall performance was also mentioned in a number of studies \cite{Koch2013, Koch2013a}. In these models, each set of chemical features was trained by specific algorithms (SVM, DT, k-NN, NB, ANN or any other algorithms) to generate one particular output. After that, these outputs were pushed to the second model which was usually another multi-layers perceptron model before giving the final result. The disadvantage of this approach is that we need to train each feature set separately because of completely different algorithms. As a result, it is difficult to build a pipeline for all training algorithms. In other words, the automation of the training procedure was reduced. 

Regarding the interpretation of neural network model, there are some interests in making neural network models more explainable and interpretable. 
A worthy approach needs to be mentioned is the use of Attention mechanisms in sequence-to-sequence model. With the encoder-decoder architecture, the attention approach not only improves the performance but reveals the alignment between input and output \cite{Bahdanau2014}.
% A worthy approach need to be mentioned is the use of \textit{Class Activation Maps} (CAM) by B. Zhou \textit{et al.} \cite{zhou2016learning} in the problem of image classification. By using the connection between the activation layer and class-specific weight, CAM method is able to visualize the region in the input which contributed to the classification. 
% Beside that, the Attention mechanisms are other methods to improve the accuracy and interpretability of sequence-to-sequence model. With the encoder-decoder architecture, the attention approach not only improves the performance but reveals the alignment between input and output \cite{Bahdanau2014}.

To deal with the problem of both automation and interpretation in predicting biological activity, we made an effort to combine different types of chemical features in one deep learning architecture and also integrate Attention mechanism. As a result, our model could train concurrently several feature sets and explain the interaction between the features and outcomes.\footnote{Our models are available at \href{https://github.com/lehgtrung/egfr-att}{https://github.com/lehgtrung/egfr-att}}

\section{Background}
\subsection{Overview of Neural network}
\subsubsection{Artificial neural network}
The Artificial neural network is computing architecture which enables a computer to learn from historical data. Nowadays, it is one of the main tools used in machine learning. As the name suggests, artificial neural networks are inspired by how biological neurons work, however,  an artificial neural network is a composition of many differentiable functions chained together. Mathematically, a neural network is a non-linear mapping which assumes the output variable $y$ as a non-linear function of its input variables $x_1,x_2,...x_n$ 
\begin{equation}
    \label{nn1}
    y = f \left(x_1,x_2,...,x_n;\theta \right) + \epsilon
\end{equation}
where $\theta$ is the parameters of the neural network and $\epsilon$ is model's inreducible error.

A very simple neural network which contains only input and output is described as follows:
\begin{equation}
    \label{nn2}
    \hat{y} = \sigma \left( w_0 + x_1 w_1 + x_2 w_2 + ... + x_n w_n \right)
\end{equation}
where $\hat{y}$ is an approximation of $y$. 

\begin{figure}[htbp]
    \centering
    \includegraphics[height=3cm]{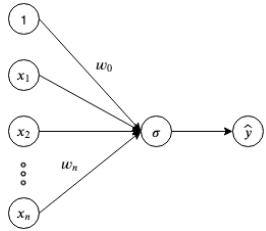}
    \caption{Example of an ordinary neuron}
    \label{fig:neuron}
    % \end{figure*}
\end{figure}

As shown in Fig. \ref{fig:neuron}, each input variable $x_i$ is represented as a node in the input layer and connects to the output node through a connection with weight $w_i$. Note that a node with value $1$ is attached to the input layer with the corresponding connection weight $w_0$ to represent the offset term. The $\sigma$ function is called \textit{activation function} or \textit{squash function} which introduce non-linear relationship between input $\mathbf{x}$ and output $y$. The \textit{Sigmoid} $\left( f(x) = \frac{1}{1+e^{-x}}\right) $ and \textit{ReLU} $\left( f(x) = max(0, x) \right)$ function are the most widely used activation functions. The model in Fig. \ref{fig:neuron} is referred as \textit{generalized linear} model. 

\begin{figure}[htbp]
    \centering
    \includegraphics[height=5cm]{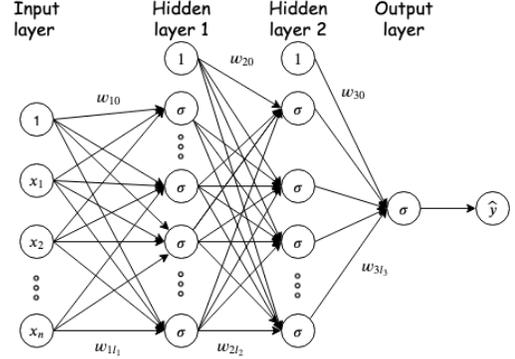}
    \caption{Example of an 2 layered neural network}
    \label{fig:neuralnet}
    % \end{figure*}
\end{figure}

The generalized linear model is simple, thus may not be able to describe complex relationship between inputs and outputs. Therefore, this architecture can be extended by building multiple generalized linear model as the form of \textit{layers} (or fully connected layers or hidden layers) and stacking those layers together to build a \textit{neural network}. Fig. \ref{fig:neuralnet} illustrates a two layered neural network, we also add neuron $1$ to the second layer as we do for the generalized linear model below. Let $l^{(k)}$ and $\hat{y}$ represent the $k$-th hidden layer and output layer respectively, neural network in Fig. \ref{fig:neuralnet} can be described mathematically as follows:
$$l_i^{(1)} = \sigma \left(w^{(1)}_{10} + \sum w^{(1)}_{1i}x_i \right) $$
$$l_i^{(2)} = \sigma \left(w^{(2)}_{20} + \sum w^{(2)}_{2i}l_i^{(1)} \right) $$
$$\hat{y} = \sigma \left(w^{(3)}_{30} + \sum w^{(3)}_{3i}l_i^{(2)} \right) $$

% Those equations can be written more compactly in matrix notation:
% $$l_1 = \sigma \left(W^{(1)T} x + \beta^{(1)} \right)$$
% $$l_2 = \sigma \left(W^{(2)T} l_1 + \beta^{(2)} \right)$$
% $$\hat{y} = \sigma \left(W^{(3)T} l_2 + \beta^{(3)} \right)$$
% where weights and bias terms from layer $k$ are stacked in a weight matrix $W^{(k)}$ and vector $\beta^{(k)}$ 

Note that a neural network may consist of an arbitrary number of many layers by simply stacking more layers. A network containing more than one layers is usually called a \textit{deep neural network}.

\subsubsection{Convolutional neural network}
Convolutional neural network (CNN) \cite{LeCun1998} is a class of neural network. The models using CNN are usually designed to operate on data with grid-like topology. CNN models are usually considered as the state-of-the-art architectures in the computer vision related tasks.
% many tasks, such as image classification, object localization, sentiment analysis [cite]. 
CNNs are also applied in biological tasks and achieved remarkable results \cite{Gawehn2016, Chen2018}. 

% \begin{figure*}[htbp]
\begin{figure}[htbp]
    \centering
    \includegraphics[width=\linewidth]{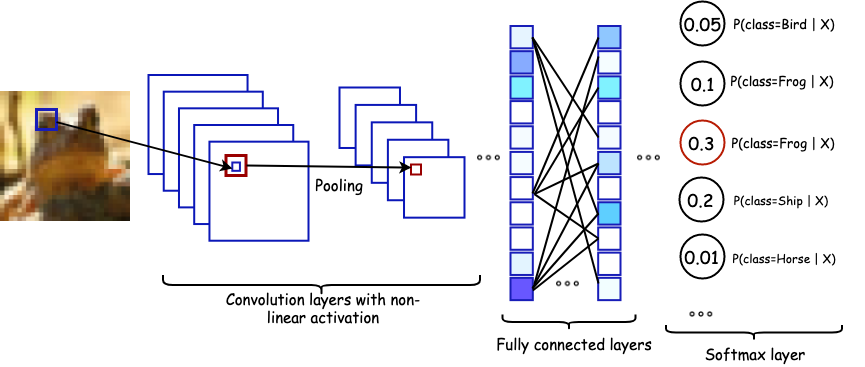}
    \caption{Example of a CNN for the Image Classification task on CIFAR10 dataset.}
    \label{fig:cnn}
    % \end{figure*}
\end{figure}

Basically, a CNN block is a combination of convolution layers followed by non-linear activation and pooling layers.
\begin{itemize}
    \item \textit{Convolutional layer (CONV)}: A convolutional layer is composed of a set of kernels to extract local features from the previous input. Each kernel is represented as a 3D tensor $F_k \in \mathbb{R}^{w \times w \times c}$, where $w$ is the size of the kernel (typically $3$ or $5$) and $c$ denotes the total number of kernels. Since $c$ is equal to the input's third dimension, it is frequently omitted when referring to the kernel shape. For an input $X \in R^{d_1 \times d_2 \times d_3}$, each kernel $F_k$ convolves with $X$ to attain a single feature map $O^k \in \mathbb{R}^{((d1-w+)/s+1) \times (d2-w)/s+1)}$ where
    \begin{equation}
        O^k_{i,j} = \sum_{i'=1}^{w}\sum_{j'=1}^{w}\sum_{l=1}^{c}[X]_{i+i'-1,j+j'-1,l}[F_k]_{i',j',l}
    \end{equation}
    where $[X]_{i,j} \in \mathbb{R}^{w\times w\times d_3}$ is a small block (known as receptive field) of $X$ around location $(i,j)$; $s$ is the stride which is the interval of the receptive fields of neighboring units.
    \item \textit{Pooling layer (POOL)}: Pooling layer creates a summary of learned features from CONV layers by aggregating the information of nearby features into a single one. The most common design of pooling operation is max-pooling. For example, a $2 \times 2$ max-pooling filter operating on a particular feature map $F$ with size $(n, m)$ will compute $\max \{F_{i,j}, F_{i+1,j}, F_{i,j+1}, F_{i+1,j+1} \}$ for each coordinate $i, j$ in $F$. This will result in a new features map with size $(m/2, n/2)$. Since a CNN typically contains multiple stacking of CONV layers, pooling is used to reduce data dimension which causes the model less computationally expensive. Pooling can also make the model invariant to small positional and translational changes.
\end{itemize}
A typical CNN architecture is generally made up of series of CNN blocks followed by one or more fully connected layers at the end. Fig. \ref{fig:cnn} illustrates a simple CNN architecture for image classification problem. 

\subsubsection{Training Neural Networks}
% During the training process, the network learns to minimized a pre-defined objective function (or loss function), which measures the discrepancy between model output $\hat{y}$ and the ground truth $y$, by adjusting it's parameters $\theta$.
The goal of learning is to minimize the \textit{loss function} with respect to the network parameters $\theta$. To do that, we need to find an estimate for the parameters $\hat{\theta}$ by solving an optimization problem of the form
\begin{equation}
    \label{thetahat}
    \hat{\theta} = \arg \min_{\theta}J(\theta)= \frac{1}{n}\sum_{i=1}^{n}\mathcal{L}(x_i, y_i; \theta)
\end{equation}
where $n$ is the number of instances in the training dataset; $\mathcal{L}$ is the loss function which measures the discrepancy between model output and the ground truth. Because the optimization problem does not have closed form solution, the method of \textit{gradient descent} is used. Firstly, the parameters $\theta$ are randomly constructed, for every iteration, the parameters are updated as follow
\begin{equation}
    \theta_{t+1} = \theta_t - \gamma\nabla_{\theta} J(\theta)
\end{equation}
 This process continues until some criterion is satisfied. Here, $\gamma$ is a constant called the \textit{learning rate} which is the amount that the weights are updated during training. As presented in the equation \ref{thetahat}, the loss function is computed over all the examples, which is computationally extensive. In practice, we use a modified version of gradient descent called \textit{stochastic gradient descent} , that means, we do not use the whole dataset for gradient computation but a subset of data called a \textit{mini-batch}. Typically, a mini-batch contains from dozens to hundreds of samples depending on system memory. Since the neural network is a composition of multiple layers, the gradient with respect to all the parameters can be methodically computed using the chain rule of differentiation also known as \textit{back-propagation} algorithm.

\subsubsection{Regularization}
One of the major issues in training neural networks is \textit{overfitting}. Overfitting happens when a network performs too well on the data it has been trained on but poorly on the test set which it has never seen before. This phenomenon is due to the large number of parameters in the network. \textit{Regularization} is able to regulate a network activity to ensure the model actually learns the underlying mapping function not just memorizing the input and output. Recently, there are two advanced regularizers which are widely used in the deep neural network.
\begin{itemize}
    \item \textit{Dropout}: During training, some weights in the network at a particular layer could be co-adapted together which may lead to overfitting. Dropout tackles this issue by randomly skipped some weights (explicitly set them zero) with a probability $p$ (usually $p=0.5$ or $0.8$). During inference, dropout is disabled and the weights are scaled with a factor of $p$ \cite{Dropout}.
    \item \textit{Batch normalization}: Recall in regression analysis, one often standardizes the designed matrix so that the features have zero mean and unit variance. This action called \textit{normalization} speeds up the convergence and make initialization easier. \textit{Batch normalization} spread this procedure to not only input layer but all of the hidden layers. During training, let $x_i$ is values across a mini-batch $\mathcal{B} = \{x_1, x_2, ..., x_k \}$, the batch norm layer calculate normalized version $\hat{x}_i$ of $x_i$ via:
    $$\hat{x}_i = \frac{x_i - \mu_{\mathcal{B}}}{\sqrt{\sigma^2_{\mathcal{B}} + \epsilon}}$$
    where $\mu_{\mathcal{B}} = \frac{1}{k}\sum_{i=1}^{k}x_i$; $\sigma^2_{\mathcal{B}} = \frac{1}{k} \sum_{i=1}^{k}(x_i - \mu_{\mathcal{B}})^2$ are mini-batch mean and variance respectively, $\epsilon$ is a constant to help computational efficiency. To make it more versatile, a batch norm layer usually has two additional learnable parameters $\gamma$ and $\beta$ which stand for scale and shift factor such that:
    $$\hat{x}_i = \gamma \hat{x}_i + \beta$$
    During inference, mini-batch mean and variance are replaced by population mean and variance which are estimated during training \cite{Batchnorm}.
\end{itemize}

\subsubsection{Attention mechanism} 
Neural networks could be considered as a "black box" optimization algorithm since we do not know what happens inside them. Attention mechanism enables us to visualize and interpret the activity of neural networks by allowing the network to look back to what it has passed through. This mechanism is motivated by how we, human, pay visual attention to certain regions of images or important words in a sentence. In the neural network, we can simulate this behavior by putting \textit{attention weights} to express the \textit{importance} of an element such as pixel in an image or a word in a sentence.

Attention mechanism was applied widely and now becomes a standard in many tasks such as Neural machine translation \cite{VaswaniSPUJGKP17}, Image captioning \cite{Anderson201}.

\subsection{Overview of EGFR}
%Introduction to egfr 
Epidermal Growth Factor Receptor (EGFR) is a member of ErbB receptor family that consists of 4 types: EGFR, HER1, HER2/new, HER3, and HER4. They are located in the cell membrane with the intrinsic tyrosine kinase. The binding of ligands (TGF-$\alpha$, amphiregulin, and other ligands) and EGFR triggers the signal amplification and diversification which lead to cell proliferation, apoptosis, tumor cell mobility, and angiogenesis. In some type of cancer (such as lung cancer), the overexpression and constitute activation cause the dysregulation of EGFR pathway that activates the tumor process \cite{Scagliotti2004, Lurje2010}. 

% The related study
For two decades, there was a great deal of effort in studying this target to discover novel medicine. 3D-QSAR was studied widely to analysis the molecular filed of ligands, which reveals the relationship between various substituents on molecules and biological activity \cite{Assefa2003, Kamath2003, Bathini2016, Zhao2017, Ruslin2019, Verma2016}. Other methods were also useful. R. Bathini \textit{et al.} employed the molecular docking and molecular mechanics with generalized born surface area (MM/GBSA) to calculate the binding affinities of protein-ligand complexes \cite{Bathini2016}. G. Verma \textit{et al.} conducted pharmacophore modeling in addition to 3D-QSAR to generate a new model which was used for screening novel inhibitors \cite{Verma2016}. 

% Machine learning study on EGFR
Regarding the application of machine learning techniques in EGFR inhibitors discovery, H. Singh \textit{et al.} \cite{Singh2015} used Random Forest algorithms to classify EGFR inhibitors and non-inhibitors. In their study, the authors collected a set of diverse chemical and their activity on EGFR. A model with high accuracy was trained and validated by 5-fold cross-validation (0.49 in MCC and 0.89 in AUC).

\subsection{Overview of Features set}
\subsubsection{SMILES Feature matrix}
SMILES (Simplified Molecular Input Line Entry System) is a way to represent the molecules in \textit{in silico} study. This method uses a strict and detailed set of rules to interpret the molecular structure into the chemical notation which is user-friendly but also machine-friendly. In particularly, SMILES notation of a molecule is a chain of character which is specified for atoms, bonds, branches, cyclic structures, disconnected structures, and aromaticity \cite{Weininger1988}.

\begin{figure}[htbp]
    \centering
    \includegraphics[width=\linewidth]{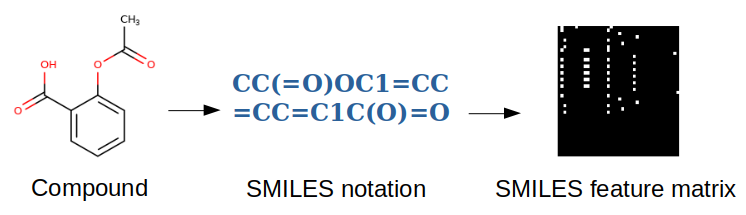}
    \caption{The steps to generate the feature matrix from chemical structure using SMILES notation}
    \label{fig:smilesmatrix}
\end{figure}

Based on this representation, M. Hirohara \textit{et al.} developed a SMILES-based feature matrix to train a convolutional neural network model for predicting toxicity. His model outperformed the conventional approach and performed comparably against the winner of Tox21 data challenge \cite{Hirohara2018}. In this dataset, each molecule was represented in the form of SMILES notation and the output consisted of 12 tasks to predict \cite{tox21}.

\subsubsection{Molecular descriptors}
Molecular descriptors are terms that characterize a specific aspect of a molecular, including substituent constants and whole molecular descriptors \cite{Of2011}. The calculation of former type derived from the difference in functional group substitution into the main core of the compound. Based on this approach, the latter is the expansion of the substituent constant. However, some whole molecular descriptors are developed from totally new methods or based on physical experiments \cite{Roy2015}.

\section{Method}
\subsection{Dataset}
The dataset used in this study was collected by H. Singh \textit{et al.} \cite{Singh2015}. This dataset contains 3492 compounds which is classified as \textit{inhibitor} or \textit{non-inhibitor} of EGFR. The inhibition activity of a particular substance was assigned if its $IC50$ is less than 10 $nM$. The ratio of inhibitors over non-inhibitor is $506:2986 \approx 1:6$. The information of chemical includes ID, SMILES representation, and class (1 for inhibitor and 0 for non-inhibitor).

% \begin{figure}[htbp]
%     \centering
%     \includegraphics[width=0.6\linewidth]{ratio}
%     \caption{The distribution of inhibitor activity in the dataset}
%     \label{fig:ratio}
% \end{figure}

\subsubsection{SMILES Feature matrix generation}
Based on the collected dataset, the chemical structure data in the form of SMILES notation was preprocessed and converted to the canonical form which is unique for each molecule by the package \texttt{rdkit} \cite{rdkit}. \\
In this study, the SMILES Feature matrix generation method developed by M. Hirohasa \textit{et al.} was used to encode the chemical notation. The maximum length of each input was 150 and thus the input strings with length below 150 were padded with zeros at the tail. In their method, for each character in the SMILES string, a 42-dimensional vector was computed (Table \ref{fttab}). The first 21 features represent the data about the atom and the last 21 features contain SMILES syntactic information \cite{Hirohara2018}.

\begin{table}[h]
\caption{Features table}
\centering
\begin{tabular}{lll}
\hline
\textbf{Features}    & \textbf{Description}                      & \textbf{Size} \\ 
\hline
Type of atom         & H, C, O, N, or others                     & 5             \\
No. of Hs            & Total number of attached Hydrogen atoms   & 1             \\
Degree               & Degree of unsaturation                    & 1             \\
Charge               & Formal charge                             & 1             \\
Valence              & Total valence                             & 1             \\
Ring                 & Included in a ring or not?                & 1             \\
Aromaticity          & Included in a aromatic ring or not?       & 1             \\
Chirality            & R, S, or others                           & 3             \\
Hybridization        & $s, sp, sp^2, sp^3, sp^3 d, sp^3 d^2,$ or others & 7             \\
SMILES symbol       & \ttfamily ( ) [ ] . : = \# \texttt{\char`\\}  / @ + -     & 21            \\
                    & \ttfamily Ion\_charge Start End         &               \\
\hline
Total                &                                           & 42            \\ 
\hline
\end{tabular}
\label{fttab}
\end{table}
\subsubsection{Descriptor calculation}
We used the package \texttt{mordred} built by H. Moriwaki, Y. Tian, N. Kawashita \textit{et al.} \cite{Moriwaki2018} to generate molecular descriptor data. Because the SMILES notation do not provide exact 3D conformation, the 2D descriptors were only calculated with total of 1613 features . The generated data was preprocessed by imputing the meaningless features or the variables which are same for whole dataset. A standard scaler was also used to normalize the molecular descriptors dataset. We used package \texttt{numpy} \cite{Oliphant2015}, \texttt{pandas} \cite{VanderWalt2011} and \texttt{scikit-learn} \cite{scikit-learn} for this process.

\subsection{Model architecture}

\begin{figure*}[htbp]
    \centering
    \includegraphics[width=0.7\linewidth]{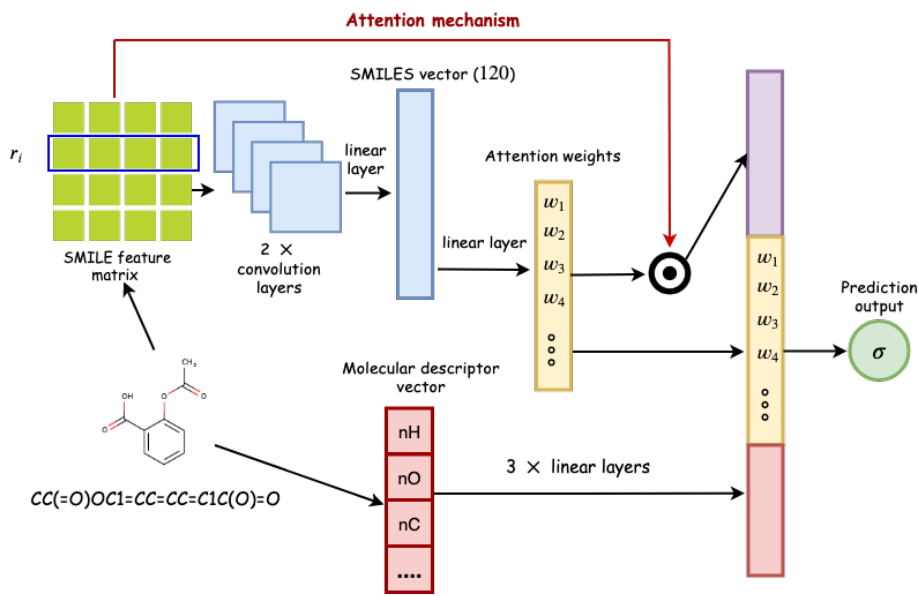}
    \caption{The Architecture of Attention based Multi-Input deep learning model}
    \label{fig:full_arch}
\end{figure*}

\subsubsection{Convolutional neural network (CNN) branch}
The SMILES Feature matrix was flown through 2 CNN blocks each consisted of a 2D convolution layer, one Batch normalization layer, a Dropout layer, and a Max pooling layer before being flattened and fully connected via a linear layer. The detail of the hyper-parameters of each layer is represented in Table \ref{cnnhp}. 

\begin{table}[htbp]
    \caption{Hyper-parameter in CNN branch}
    \begin{center}
        \begin{tabular}{llc}
            \hline
            \textbf{Layer}                               & \textbf{Hyper-parameter}          & \textbf{Value} \\
            \hline
            $1^{st}$ conv2d                     & No. of input channels  & 1     \\
                                                & No. of output channels & 6     \\
                                                & Kernel size            & (3,3) \\
                                                &                         &         \\
            $2^{nd}$ conv2d                     & No. of input channels  & 6     \\
                                                & No. of output channels & 16    \\
                                                & Kernel size            & (3,3) \\
                                                &                         &         \\
            Dropout                             & Dropout rate           & Be tuned   \\
            \hline
        \end{tabular}
        \label{cnnhp}
    \end{center}
\end{table}

\begin{table}[htbp]
    \caption{Hyper-parameter in MD branch}
    \begin{center}
        \begin{tabular}{llc}
            \hline
            \textbf{Layer}                               & \textbf{Hyper-parameter} & \textbf{Value} \\
            \hline
            {$1^{st}$ Linear layer}    & No. of neurons              & 512     \\
            {$2^{nd}$ Linear layer}    & No. of neurons              & 128     \\
            {$3^{rd}$ Linear layer}    & No. of neurons              & 64     \\
            \hline
        \end{tabular}
        \label{mdhp}
    \end{center}
\end{table}

\subsubsection{Attention mechanism}

The idea of using the attention mechanism came from the fact that each chemical's atoms contribute differently to the drug's effect. In other words, we put \textit{attention} or \textit{weight} to the atoms in the chemical which are represented by rows in the SMILES feature matrix. The larger the weight of an atom is, the more contribution of which atom contribute to the drug. By doing this, we can extract each components weights for interpreting the results and analysis.

Let $\Vec{m}$ and $R$ denote the vector obtained by feeding SMILE vector through a linear layer and the SMILE feature matrix,  respectively. The model uses the similarity between $\Vec{m}$ and the $i$-th row $\Vec{R}_i$ as a measure of the importance of $\Vec{R}_i$. Particularly, let $\Vec{a}_i$ denote the \textit{attention weight} vector, it is formulated as follows:
\begin{equation}
    \Vec{a}_i = \frac{1}{1 + \exp \left( \Vec{R}_i \cdot \Vec{m} \right)}
\end{equation}

After that, the vector $\Vec{a}$ is then used as the coefficient of a linear combination of rows in the SMILES feature matrix. Therefore, the output of the attention layer $\Vec{f}$ is expressed as:
\begin{equation}
    \Vec{f} = \sum \Vec{a}_i \times \Vec{R}_i
\end{equation}

% The vector obtained from the linear combination is concatenated with the attention weight vector and then was fed through a linear layer and squashed by sigmoid activation function to make a prediction as in Fig. \ref{fig:arch}.

\subsubsection{Molecular Descriptors (MD) branch}
The MD branch is used to train the molecular descriptors data. Let $\Vec{t}$ is the vector obtained by feeding molecular descriptors vector through 3 blocks of fully connected layers each consisted of a fully connected layer, a batch normalization layer, and a dropout layer. $\Vec{t}$ is considered as a high-level representation of molecular descriptors data. 

\subsubsection{Concatenation}
The three vectors: $\Vec{f}$, $\Vec{m}$, $\Vec{t}$ are then concatenated as in Fig. \ref{fig:full_arch} to form the final representation vector. This vector combines information from both molecule structural information and descriptors information. It is then fed through a linear layer and squash by sigmoid function to make the final prediction as the probability of the molecule as an inhibitor.

\subsection{Hyper-parameter tuning}
In this study, \texttt{PyTorch} platform \cite{paszke2017} was used in order to implement our model and the 5-fold cross-validation was conducted to evaluate the performance. In this method, the dataset was split into 5 parts. For each fold, the model was trained on the set of 4 parts and tested on the remaining part. The choice of the training set was permuted through all divided parts of the dataset, thus the model was trained 5 times and the average performance metrics of each time was used to evaluate. The ending point of Training step was determined by Early-stopping technique \cite{Finnoff1993, Prechelt1998}. Thus, for each fold, the model would stop training if the loss value increases continuously 30 epochs. 

The second column of Table \ref{hpt} presents the hyper-parameters and their considered values in the tuning step. Grid Search technique was conducted to determine the best combination of hyperparameters which had the best performance. However, in the case of discovering the suitable batch size for training, several suggested values were tested and the chosen was the value which utilized the system efficiently. Additionally, the threshold of the classifier was determined by analyzing the ROC plot and the Precision-Recall curve. The most optimal threshold was the point nearest to top-left of the ROC plot and gave the balance between Precision and Recall in the latter plot.

\begin{table}[htbp]
    \caption{Hyper-parameter tuning}
    \begin{center}
        \begin{tabular}{lcc}
            \hline
            \textbf{Hyper-parameter}    & \textbf{Value}     & \textbf{Optimal value}\\
            \hline
            Batch size                  & 32; 64; 128; 512  & 128                   \\
            Dropout rate                & 0; 0.2; 0.5         & 0.5                     \\
            Optimizer                   & SGD; ADAM         & ADAM                  \\
            Learning rate                & 1e-4; 1e-5; 1e-6     & 1e-5                    \\
            Threshold                    & 0.2; 0.5; 0.8        & 0.2                    \\
            %Epochs                        & 100...1000        & 600                    \\
            \hline
        \end{tabular}
        \label{hpt}
    \end{center}
\end{table}

\subsection{Performance Evaluation}
In order to assess the performance of each model, several metrics were calculated during training and validation steps (Table \ref{metrics}).
%\begin{itemize}
%    \setlength\itemsep{0.5em}
%    \item \makebox[2cm]{$\text{Sensitivity}$\hfill}  $= \frac{\text{TP}}{\text{TP} + \text{FN}}$
%    \item \makebox[2cm]{$\text{Specificity}$\hfill} $= \frac{\text{TN}}{\text{TN} + \text{FP}}$
%    \item \makebox[2cm]{$\text{Accuracy}$\hfill} $= \frac{\text{TP}+\text{TN}}{\text{TP}+\text{TN}+\text{FP}+\text{FN}}$
%    \item \makebox[2cm]{MCC\hfill} $=\frac{\text{TP} \times \text{TN} - \text{FP} \times \text{FN}}{\sqrt{(\text{TP}+\text{FP})(\text{TP}+\text{FN})(\text{TN}+\text{FP})(\text{TN}+\text{FN})}}$
%    \item \makebox[2cm]{AUC\hfill} : the area under the ROC curve
%\end{itemize}

\begin{table}[htbp]
    \centering
    \begin{threeparttable}
        \caption{Performance Metrics}
        \label{metrics}
        
        \begin{tabular}{lc}
            \hline
            \textbf{Metrics}        & \textbf{Formulas\tnote{\textdagger}} \\
            \hline
            Sensitivity    & $\frac{\text{TP}}{\text{TP} + \text{FN}}$\\
            Specificity    & $\frac{\text{TN}}{\text{TN} + \text{FP}}$ \\
            Accuracy    & $\frac{\text{TP}+\text{TN}}{\text{TP}+\text{TN}+\text{FP}+\text{FN}}$\\
            MCC            & $\frac{\text{TP} \times \text{TN} - \text{FP} \times \text{FN}}{\sqrt{(\text{TP}+\text{FP})(\text{TP}+\text{FN})(\text{TN}+\text{FP})(\text{TN}+\text{FN})}}$\\
            AUC            & \mbox{the area under the ROC curve} \\
            \hline
        \end{tabular}
        \begin{tablenotes}
            \footnotesize
            \item[\textdagger] TP: True Positive, FN: False Negative, TN: True Negative, FP: False Positive, MCC: The Matthews correlation coefficient.
        \end{tablenotes}
    \end{threeparttable}
\end{table}

The ROC analysis and AUC are usually considered as the most popular metrics for imbalanced dataset because they are not biased against the minor label \cite{Kotsiantis2006, Guo2009}. However, these metrics show the overall performance in the whole domain of threshold. In other words, ROC or AUC are not represented for a particular classifier. In our study, MCC was the most preferred criteria to evaluate model performance. This is because MCC considers all classes in the confusion matrix whereas other metrics (eg, \textit{accuracy} or \textit{F1-score}) do not fully use four classes in the confusion matrix \cite{Chicco2017}. The remaining metrics were still useful for the benchmark.

\section{Result}
\subsection{Hyper-parameters optimization}
Despite of imbalanced classes, the loss function of model (binary cross-entropy) still converged with the loss value of $0.10 - 0.16$ for Training set and $0.22 -0.28$ for Validation set at the end of each fold when cross-validating the model CNN + MD + ATT %(Figure \ref{fig:loss}).

% \begin{figure}[h]
%     \centering
%     \includegraphics[width=\linewidth]{loss.png}
%     \caption{Loss function during Training step}
%     \label{fig:loss}
% \end{figure}

The training batch size is 128 which gave the best utility on the GPU Tesla K80. When comparing the effect of two types of the optimizer, we observed that Adaptive Moment Estimation (ADAM) showed a better result than Stochastic gradient descent (SGD) in both running time and model performance. 

The optimal collection of hyper-parameters was listed in the third column of Table \ref{hpt}. These values were used for evaluating the performance of three considered models.

\subsection{Performance}
% Our architecture was trained on the EGFR dataset by using cross-validation method. When the model used structural information in the CNN branch only, the model performance was poor and this was even worse than H. Singh \textit{et al.} study. However, the CNN + MD model was outperforming the reference and other models. Furthermore, the attention-integrated model had slightly weaker performance but still better than referring model. In spite of lower indicator, this model had a huge advantage which helped interpret results as mentioned later in \ref{subsec:res_att}).
\begin{figure*}[htbp]
    \centering
    \includegraphics[width=\linewidth]{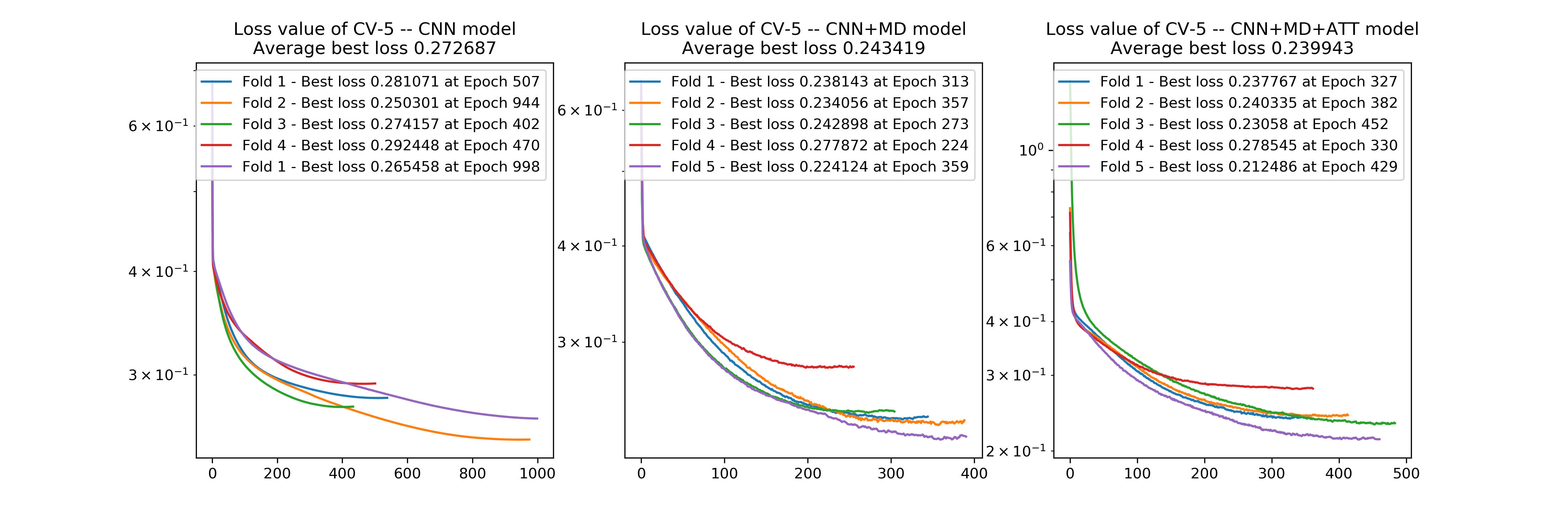}
    \caption{The Loss value of each model on the Validation steps}
    \label{fig:arch}
\end{figure*}

Our architecture was trained on the EGFR dataset and evaluated by mentioned cross-validation method. The final result represents in Table \ref{tab:perform}. When using structure information in the CNN branch only, the performance was slightly better than H. Singh \textit{et al.} model in all metrics excepting AUC. However, the CNN + MD model and CNN + MD + ATT model was outperforming to the reference model. By comparing two important indicators, it can be seen that there was an improvement in MCC and AUC. Additionally, when training with more branch (CNN + MD and CNN + MD + ATT), the Running time was also reduced significantly to around a half.

\begin{table}[htbp]    
    \centering    
    \begin{threeparttable}
        \caption{Performance Comparison }
        \label{tab:perform}
        \centering    
        \begin{tabular}{lcccc}
        \hline
        \textbf{Metrics} \tnote{\textdagger}
                        & H. Singh  & CNN   & CNN + MD & CNN + MD \\
                        &\textit{et al.}    &   &   & + ATT \tnote{\textdaggerdbl}  \\
        \hline
        \textbf{LOSS}   & N.A                      & 0.2727  & 0.2434   & 0.2399   \\
        \textbf{SENS}   & 69.89\%                  & 74.31\% & 75.29\%  & 74.11\%   \\
        \textbf{SPEC}   & 86.03\%                  & 85.77\% & 89.75\%  & 90.15\%   \\
        \textbf{ACC}    & 83.66\%                  & 84.10\% & 87.66\%  & 87.83\%   \\
        %\textbf{MCC}    & 49.00\%                  & 49.90\%  & 55.49\%  & 55.85\%  \\
        \textbf{MCC}    & 0.49                  & 0.50  & 0.58  & 0.57  \\
        \textbf{AUC}    & 89.00\%                  & 87.52\%  & 90.32\%  & 90.84\%  \\
        %\textbf{RT}     & N.A                      & 42 min   & 17 min   & 22 min   \\
        \hline      
        \end{tabular}
        \begin{tablenotes}
            \footnotesize
            \item[\textdagger] LOSS: Average Best Loss value in 5-fold Cross validation, SENS: Sensitivity, SPEC: Specificity, ACC: Accuracy, MCC: The Matthews correlation coefficient, AUC: The Area under the ROC curve, RT: Running time.
            \item[\textdaggerdbl] CNN: using CNN branch only, CNN + MD: using both CNN and MD branch, CNN + MD + ATT: using both CNN and MD branch with Attention mechanism.
        \end{tablenotes}
    \end{threeparttable}
\end{table}

% CNN + MD + ATT: https://www.floydhub.com/kiengcan9999/projects/egfr-cam/110
% CNN + MD      : https://www.floydhub.com/kiengcan9999/projects/egfr-cam/112
% CNN           : https://www.floydhub.com/kiengcan9999/projects/egfr-cam/114

% Regarding \textit{AUC}, there is also a slight improvement. As mentioned previously, this metric was not the first criteria to benchmark. The significant enhancement in \textit{MCC} was the clear evidence of the advantage of our architecture comparing to the referring study (H. Singh \textit{et al.} \cite{Singh2015})

\subsection{Attention mechanism}\label{subsec:res_att}
\begin{figure}[ht]
    \centering
    \includegraphics[width=\linewidth]{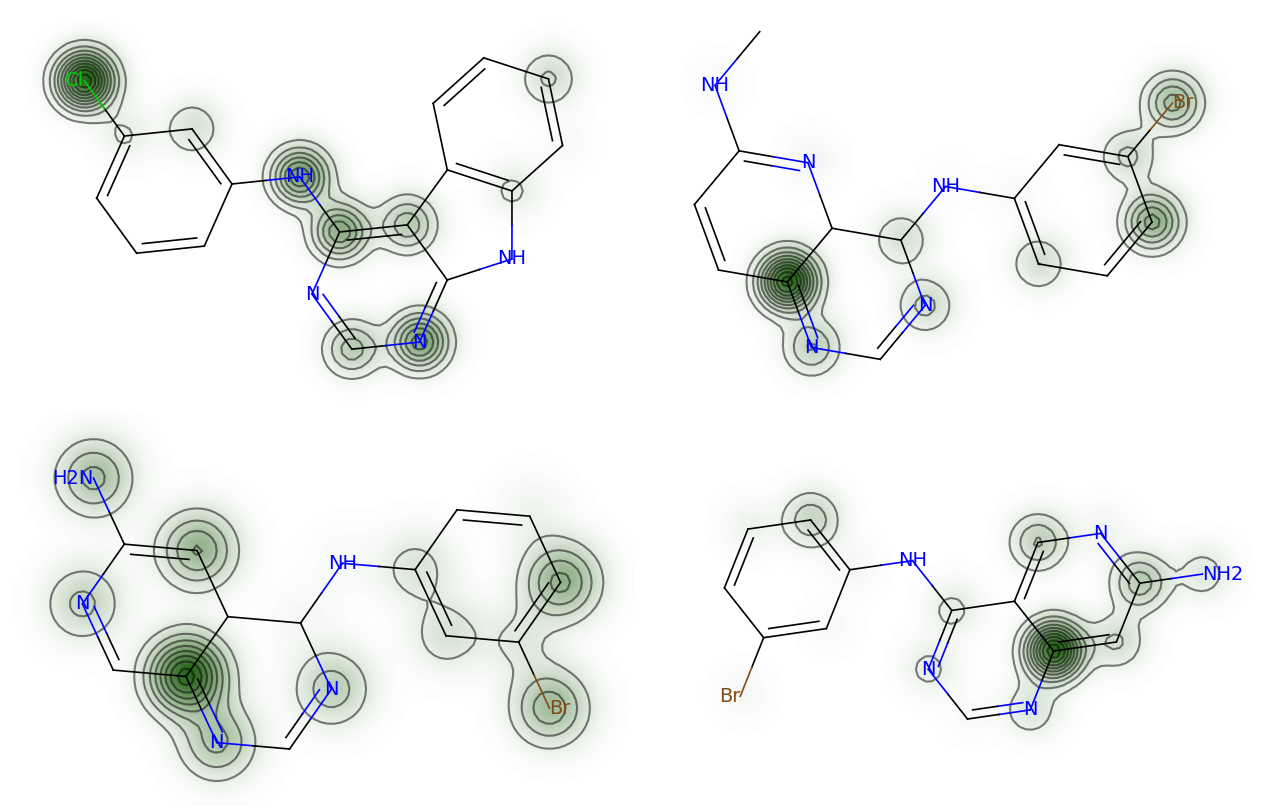}
    \caption{Chemical structure interpretation using Attention weight}
    \label{fig:weight_vis}
\end{figure}

The Attention mechanism was successfully implemented in our architecture. From the attention weights vector, the weights representing to each atom in the molecules were extracted and used to indicate their contribution to the compound's activity. By using visualization on the package \texttt{rdkit}, these attention weights could be used to visualize the distribution of contribution over the molecular structure. Figure \ref{fig:weight_vis} illustrates some example from the model.

Additionally, the integration of attention mechanism to the model did not give much more improvement in performance. We believed that this problem could be improved by optimizing the hyperparameters. However, we did not focus on hyperparameters optimization for this model because the aim of Attention vector was to make the model more interpretable.

The visualization of Attention vectors revealed some key findings in chemical structure, such as the Nitrogens in the hetero-cyclic structure are usually highlighted by the model and the halobenzyl substitution on heterocyclic structure contributed positively to the bioactivity.

\section{Discussion}
There are two major advantages to our architecture. The first strong point is the combination of both structure information and chemical attributes in a single learning model. As a result, this advancement made a significant improvement in both performance and automation. Another worthy innovation was the integration of attention mechanism which facilitated the interpretation of the model. In fact, attention weight generated by the model would help explain the contribution of each atom on the overall biological activity.

Comparing to another effort to make deep learning model more interpretable, our architecture has an advantage in computation because it is easier to generate the SMILES Feature matrix than other algorithms. For example, Sanjoy Dey \textit{et al.} \cite{Dey2018} used ECFP fingerprint algorithms to transform the chemical structure into the matrix feature. This method does not treat the molecule as a whole structure but calculate on each fragment of a chemical with a particular radius. Additionally, there are required calculation to generate the features including tuning the hyper-parameters of the algorithms (e.g the radius of calculation).

Regarding to our implementation of Attention mechanism, we observed that each atom in a substance was treated separately; as a result, the connection between atom was not highlighted in our model, as well as the contribution of some functional groups which contain many atoms (e.g, carbonyl, carboxylic, etc) was not clearly illustrated. We proposed a solution for this limitation that is to add another branch to the architecture which embeds the substructure patterns (e.g, Extended-Connectivity Fingerprints \cite{Rogers2010}, Chemical Hashed Fingerprint \cite{Al-Lazikani2004}).

Additionally, the lower performance of the CNN model comparing to the baseline model was another interesting finding. This could be due to the sparsity of SMILES feature matrix. In fact, the CNN as well as other deep learning algorithms require much data to accumulate the information in the training step. In case of SMILES feature data, because of zero paddings to justify the length of encoding vectors, the feature matrix became sparse. This led to the fact that the model required more data for training but the dataset was quite small for deep learning. However, in the case of CNN + MD or CNN + MD + ATT model, because of the addition of another input data, the models acquired information more easily. As a result, the performance was improved in terms of all metrics.

When considering the running time between different models, it is clear that the longest running time was that of model with only CNN branch (59 min) while the more complicated model with more data like CNN + MD and CNN + MD + ATT took just a half of running time with 31 min and 37 min, respectively. This could be because the SMILES Feature matrix in the CNN model was sparse so the model should train longer to achieve the convergence of loss function. In the CNN + MD and CNN + MD + ATT model, there could be a complement between different input branches and we supposed that there was an information flow transferring between two branches, which facilitated the training stage and performance improvement. In other studies which also used several types of data \cite{Koch2013, Koch2013a}, the model trained model separately and did not use this information connection. This phenomenon might represent an advantage of our architecture.

In conclusion, the combination of different source of features is definitely useful for bioactivity prediction, especially when using deep learning model. The attention-based multi-input architecture we proposed achieved a superior score comparing to referring model. Additionally, the attention mechanism would help to interpret the interaction between each element of chemical structures and their activity.

\bibliographystyle{ieeetr}
\bibliography{paper}

\end{document}